\documentclass[runningheads]{llncs}
\usepackage[T2A,T1]{fontenc}
\usepackage[russian,english]{babel}
\usepackage{graphicx}
\usepackage{paralist}
\usepackage{hyperref}
\usepackage{xspace}
\usepackage{float}
\usepackage[inkscapelatex=false]{svg}
\usepackage{setspace}
\usepackage{anyfontsize}

\newcommand{\cyrill}[1]{{\foreignlanguage{russian}{#1%
}}}

\begin{document}
\selectlanguage{english}

\title{KyrgyzNLP: Challenges, Progress, and Future}
\author{Anton Alekseev\inst{1,2,3,4}\orcidID{0000-0001-6456-3329} \and Timur~Turatali\inst{5}\orcidID{0009-0001-6778-460X}}
\authorrunning{A. Alekseev, T. Turatali}

\institute{
Steklov Mathematical Institute at St. Petersburg, St. Petersburg, Russia \and
St. Petersburg State University,  St. Petersburg, Russia \and
Kazan (Volga Region) Federal University, Kazan, Russia \and
Kyrgyz State Technical University n. a. I. Razzakov, Bishkek, Kyrgyzstan \and
The Cramer Project, Bishkek, Kyrgyzstan \and
Kyrgyz AI Research Institute, Bishkek, Kyrgyzstan\\
\email{\{anton.m.alexeyev,timur.turat\}@gmail.com}}

\maketitle
\begin{abstract}
Large language models (LLMs) have excelled in numerous benchmarks, advancing AI applications in both linguistic and non-linguistic tasks. However, this has primarily benefited well-resourced languages, leaving less-resourced ones (LRLs) at a disadvantage. In this paper, we highlight the current state of the NLP field in the specific LRL: \cyrill{кыргыз тили}.

Human evaluation, including annotated datasets created by native speakers, remains an irreplaceable component of reliable NLP performance, especially for LRLs where automatic evaluations can fall short. In recent assessments of the resources for Turkic languages, Kyrgyz is labeled with the status ``Scraping By'', a severely under-resourced language spoken by millions. This is concerning given the growing importance of the language, not only in Kyrgyzstan but also among diaspora communities where it holds no official status.

We review prior efforts in the field, noting that many of the publicly available resources have only recently been developed, with few exceptions beyond dictionaries (the processed data used for the analysis is presented at~\url{https://kyrgyznlp.github.io/}). While recent papers have made some headway, much more remains to be done. Despite interest and support from both business and government sectors in the Kyrgyz Republic, the situation for Kyrgyz language resources remains challenging. We stress the importance of community-driven efforts to build these resources, ensuring the future advancement sustainability. We then share our view of the most pressing challenges in Kyrgyz NLP. Finally, we propose a roadmap for future development in terms of research topics and language resources.

\end{abstract}


\section{Introduction}\label{sec:introduction}

In recent years, Natural Language Processing (NLP) and Large Language Models (LLMs) have revolutionized the field of artificial intelligence, pushing the boundaries of what machines can accomplish in understanding and generating human language. These advancements have led to significant improvements in various applications, from machine translation to sentiment analysis, and from chatbots to content generation. What might be even more important, LLMs are already changing the job markets and other spheres of life~\cite{ai_jobs,osullivan_2024}. 
Since 2022, texts written in natural languages have become a \textit{crucial medium between humans and AI applications}, making NLP arguably more prominent than ever.

However, the rapid progress in NLP has not been uniform across all languages, creating a digital divide that threatens to worsen existing linguistic inequalities.

While languages with abundant digital resources, such as English or Spanish, have reaped the benefits of these technological breakthroughs, less-resourced languages (LRLs) have largely been left behind. This disparity is particularly concerning as it risks further marginalizing communities that already face challenges in the digital age. One such language that exemplifies this issue is Kyrgyz (\cyrill{кыргыз тили}), a Turkic language spoken by millions of people primarily in Kyrgyzstan and neighboring regions.

Kyrgyz, despite its rich cultural heritage and significant speaker population, has been classified as ``Scraping By'' in terms of NLP resources~\cite{mirzakhalov2021large,veitsman2024recentadvancementschallengesturkic}. This classification underscores the severe lack of digital tools, datasets, and models available for processing and analyzing the language. The scarcity of resources is particularly troubling given the growing importance of Kyrgyz, not only within Kyrgyzstan where it holds official status but also among diaspora communities worldwide where the language serves as a vital link to cultural identity.

The development of NLP tools for less-resourced languages like Kyrgyz presents unique challenges. Unlike well-resourced languages that can rely on vast amounts of digital text for training models, LRLs often require more careful curation of limited available data. Moreover, the creation of high-quality, annotated datasets --- a cornerstone of modern NLP --- demands the involvement of native speakers and linguistic experts, a~\textit{resource} that can be scarce for many LRLs.

Recent years have seen a growing awareness of the importance of linguistic diversity in the digital sphere. Both governmental and private sector entities in Kyrgyzstan, where Kyrgyz is the official language, have shown interest in developing language technologies. 

Government initiatives include, for example,
\begin{inparaenum}[(1)]
    \item {Constitutional Law on the State Language}: on July 17, 2023, President Sadyr Japarov signed the Constitutional Law on the State Language, aiming to implement the Article 13 of the constitution, which designates Kyrgyz as the state language; this law outlines procedures for its use, underscoring the government's commitment to enhancing the status and usage of the Kyrgyz language~\cite{japarov_2023};
    \item {the National Commission on the State Language and Language Policy} operating under the President of the Kyrgyz Republic and focusing on the development and implementation of language policies that promote the Kyrgyz language; it serves as a central body for coordinating efforts related to language development and policy enforcement\footnote{\url{https://mamtil.gov.kg/en}}.
\end{inparaenum}

Private sector initiatives in Kyrgyzstan include
\begin{inparaenum}[(1)]
    \item the AI-related efforts involving Kyrgyz language processing: in early 2024, Kyrgyzstan tested an artificial intelligence model capable of processing and understanding the Kyrgyz language; this development highlights the private sector's role in advancing language technologies that support linguistic diversity\footnote{\texttt{\href{https://central-asia.media/26842-kyrgyzstan-has-tested-ai-model-in-kyrgyz-language.html}{central-asia.media:Kyrgyzstan has Tested AI Model in Kyrgyz Language.}}};
    \item collaborations to bridge the AI language gap: in May 2024, telecom company Veon, along with partners including Beeline Kazakhstan and the Barcelona Supercomputing Center, announced efforts to address the ``AI language gap'' for under-represented languages; this initiative focuses on developing tools and documentation for languages like Kyrgyz, aiming to enhance their representation in AI models\footnote{\href{https://www.reuters.com/technology/veon-teams-up-with-partners-bridge-online-ai-language-gap-2024-05-15/}{\tt Reuters:Veon teams up with partners to bridge online 'AI language gap'.}}.   
\end{inparaenum}

These examples demonstrate a concerted effort by both governmental and private entities in Kyrgyzstan to develop language technologies that promote linguistic diversity in the digital realm.

This paper aims to address the current state of Kyrgyz language processing and chart a possible course for its future development. We begin by reviewing the existing efforts in Kyrgyz NLP, noting the recent emergence of some publicly available resources\footnote{We have developed an interactive website presenting and organizing the data used in this study: \url{https://kyrgyznlp.github.io/}}. We then identify the most pressing challenges facing the field, drawing on insights from both academic research and practical applications. Finally, we propose a roadmap for future development, outlining key research areas and necessary language resources.

Our intention is to encourage a community-driven approach to resource development and to motivate the interested parties in government and business in the support of the process, to ensure sustainable progress in making advanced language technologies accessible to all linguistic communities.

The remainder of this paper is structured as follows.
Section~\ref{sec:data_nlp} discusses the importance of data in NLP, beginning with motivating examples and exploring processing methods specifically suited for less-resourced languages. It further examines the potential of large language models (LLMs) as a universal solution in this context. Section~\ref{sec:kyrgyz_lrl} introduces Kyrgyz as a less-resourced language, offering an overview of the language and its unique linguistic relations within the Turkic language family. Section~\ref{sec:challenges} addresses the challenges facing Kyrgyz NLP, including resource scarcity, script and dialect diversity, the complexities of agglutinative morphology, and the decentralized nature of existing initiatives. Then, in Section~\ref{sec:existing_surveys}, the existing surveys of the field are reviewed. Section~\ref{sec:scientometrics} presents a scientometric analysis of Kyrgyz NLP, categorizing key topics within the field and highlighting noteworthy observations. Section~\ref{sec:overview_kyrgyznlp} provides an overview of recent Kyrgyz NLP research, with a focus on morphology, syntax, semantics, corpus studies, and NLP applications.
In Section~\ref{sec:available_resources}, we review existing resources relevant to Kyrgyz NLP. Section~\ref{sec:untapped} identifies untapped subdomains of Kyrgyz NLP, emphasizing areas for new language resources and innovative, untried methodological approaches.
Section~\ref{sec:roadmap} outlines a roadmap for future development in Kyrgyz NLP, detailing proposed milestones and a projected timeline. Finally, Section~\ref{sec:conclusion} concludes the paper with a summary of key findings and implications for the continued advancement of Kyrgyz NLP.


\section{On the Importance of Data in NLP}\label{sec:data_nlp}

Data is undoubtedly the cornerstone of NLP. Carefully curated datasets have historically driven significant advancements in the field. In this section, we support this claim and clarify the reasons behind our advocacy for data annotation initiatives.

\subsection{Motivating Examples}\label{ssec:motivating_examples}
Two notable NLP milestones illustrate the impact of rigorous data preparation on the progress of language science.

\textbf{Penn Treebank.}
Released in the 1990s, the Penn Treebank~\cite{marcus1993building} introduced syntactic and part-of-speech annotations for millions of English words. This resource shifted NLP from rule-based to probabilistic, data-driven approaches, enabling breakthroughs in syntactic parsing, machine translation, and statistical modeling.

\textbf{BioBERT.} Published in the 2020s, BioBERT~\cite{lee2020biobert} leveraged annotated biomedical texts to outperform general NLP models in domain-specific tasks like named entity recognition, relation extraction, and others~\cite{miftahutdinov2019kfu,sakhovskiy2021kfu}. This demonstrated the importance of specialized data and triggered a trend in fine-tuning models on annotated, domain-specific datasets.

A more recent study~\cite{samuel2023trained} showed that training BERT~\cite{devlin2019bert} on the \emph{British National Corpus}~\cite{bnc2007} (i.e., a carefully curated yet much smaller text collection than that used to train the original model) achieved even better performance than the original BERT model.

Thus, even in the age of powerful neural models, the quality and specificity of data remain critical factors in achieving high-performance NLP systems. Moreover, without datasets for training and validation, the field of Kyrgyz NLP simply cannot advance.
\subsection{Processing Methods for Less-Resourced Languages}

Addressing the challenges faced by LRLs requires innovative approaches that compensate for the lack of resources. Several common methods have been employed to process LRLs beyond Kyrgyz.

\textbf{Data Collection and Augmentation.}
Collecting additional data is a straightforward but resource-intensive approach. This involves gathering texts from various domains, including literature, social media, and transcriptions of spoken language. Data augmentation techniques, such as back-translation~\cite{sennrich2016improving,hoang2018iterative} and synthetic data generation~\cite{lample2018phrase,rahimi2019massively}, can increase dataset size and diversity.

\textbf{Leveraging Machine Translation and Multilingual Models.}
Utilizing machine translation and multilingual models allows for transfer learning from high-resource languages. Models pre-trained on languages with similar linguistic features can be fine-tuned for Kyrgyz. For Turkic languages, transfer learning from Turkish or Kazakh can be beneficial due to linguistic similarities~\cite{mirzakhalov2021large,mirzakhalov2021evaluating}.

\textbf{Sampling and Transfer Learning.}
Sampling strategies help select the most informative data subsets for training. Transfer learning involves adapting models trained on large datasets in other languages to Kyrgyz, adjusting the model’s parameters to better fit the target language.

\textbf{Manual and Unsupervised Techniques.}
Rule-based systems, developed with linguistic expertise, can address specific challenges in morphology and syntax. Unsupervised learning methods, such as clustering and topic modeling, can extract patterns from unannotated data, which is especially useful when labeled data is scarce.

\textbf{Utilizing External Tools.}
Existing tools and frameworks, such as Apertium~\cite{forcada-tyers-2016-apertium}, a platform for rule-based machine translation, can be adapted for Kyrgyz~\cite{washington2012finite}. These tools often support multiple languages and can be extended to include LRLs with additional development.

\subsection{LLMs as a “Universal Solution”?}

An emerging counter-argument to developing datasets for low-resource languages (LRLs) and traditional non-LLM NLP methods suggests that next-generation large language models may surpass prior approaches across various NLP tasks, even without extensive task-specific data for training or fine-tuning. While an in-depth discussion of the limitations of this perspective is beyond our scope, it warrants careful consideration due to its implications for LRL data development.

A recent example involving the \emph{Gemini 1.5 Pro} model’s in-context learning is indeed impressive: \textit{When given a grammar manual for Kalamang, a language with fewer than 200 speakers worldwide, the model learns to translate English to Kalamang at a level comparable to a person learning from the same content}~\cite{google2024gemini,Visser2022,tanzer2024a}. Beyond this specific case, more approaches of this sort are likely to emerge in the future~\cite{zhang2024hire}.

Essentially, in 2024, given a sufficiently large and comprehensive prompt derived from a textbook, a proprietary model — available via API as of October 2024 — is capable of providing translations to and from previously “unseen” languages, achieving a quality level comparable to that of a human learning the same language. This is not the only example of the success of large language models (LLMs) in tasks for which, as asserted, they had no direct data in the training set.

However, we believe this progress should not demotivate those working with low-resource languages (LRLs), nor those supporting, funding, or volunteering in data annotation for related research. On the contrary, it should encourage further exploration, and the example involving the Kalamang language is particularly inspiring. Evaluation datasets can be viewed as analogous to a \textit{textbook} — though they may not contain explicit rules, they are nonetheless rooted in authentic data.

Beyond the arguments above and the examples in Section~\ref{ssec:motivating_examples}, there are many practical reasons to continue developing language resources despite the advances of universal models.

\paragraph{Safety, Decision-Making, and Model Evaluation.} All machine learning (ML) and artificial intelligence (AI) solutions, including those based on LLMs, require rigorous quality and safety control by human experts. Incorrect decisions about which foundational model to select for a given task — whether in industry or academia — can lead to unnecessary expenses, ecological impacts, security vulnerabilities, reputational harm, and other negative outcomes~\cite{kraft_2016}. Evaluating models on tasks similar to those they are expected to solve in production or at scale is arguably the most cost-effective approach to understanding their potential performance.

Further, this perspective underscores the need to create diverse and realistic datasets for model evaluation, ensuring that models are tested in scenarios that closely mirror real-world use. This approach will not only enhance model reliability but also provide an essential benchmark for comparing performance across diverse tasks, ultimately driving the development of safer and more effective AI systems for both high-resource and low-resource language contexts.

\paragraph{Cultural Preservation.} Less-resourced languages preserve unique cultures and histories. Even independent of the “new AI spring,” it is essential to document Kyrgyz language patterns across various contexts and media: news reporting, literary fiction, conversational speech, folklore, children’s language, and social media commentary. Dialectal forms and linguistic phenomena such as code-switching, neologisms, irony, and context-specific expressions demand attention, best captured through contemporary sources and spoken language. To preserve both ``classical'' and dynamic, modern expressions of Kyrgyz, it is crucial to collect, annotate, and organize text corpora. With the anticipated widespread adoption of AI solutions, there is limited time to accomplish this.

\paragraph{Innovation and Research.} Addressing the challenges posed by LRLs, such as limited data availability, complex morphology, and unique linguistic structures, fosters the development of novel methodologies that benefit the \textit{entire field of NLP}. For instance, strategies developed for low-resource settings often contribute to advancements in transfer learning, data augmentation, and model robustness — areas critical to making LLMs more adaptable and efficient across diverse linguistic contexts. By investing in resources for LRLs like Kyrgyz, the NLP field not only promotes linguistic inclusivity but also drives research with widespread implications, strengthening the adaptability and generalizability of AI across languages and domains.

\paragraph{Human Oversight.} Human oversight remains essential across NLP tasks, especially for handling \textit{edge cases and cultural nuances} that LLMs may overlook or misinterpret. While LLMs have shown impressive capabilities, their limitations in understanding context-specific subtleties, idiomatic expressions, and culturally embedded meanings highlight the need for human review. This is particularly critical for languages like Kyrgyz, where unique linguistic and cultural elements may be underrepresented in training data. Human expertise ensures that outputs are accurate, culturally sensitive, and contextually relevant, safeguarding against potential misinterpretations and biases. Human involvement is crucial not only for the safety and reliability of NLP applications but also for maintaining the integrity and richness of underrepresented languages.

\paragraph{Model Limitations.} The success of certain models, such as Gemini 1.5 in translating the Kalamang language, does not guarantee comparable performance across all languages. Additionally, while impressive, \textit{translation} alone does not cover the full spectrum of linguistic tasks. Generalization remains a challenge, as language-specific complexities often reveal gaps in performance, underscoring the need for dedicated resources and models tuned to each language’s unique characteristics.


\section{Kyrgyz as a Less-Resourced Language}\label{sec:kyrgyz_lrl}

\subsection{The Kyrgyz Language: An Overview}\label{ssec:kyrgyz101}

\paragraph{Linguistic and Demographic Profile.}

Kyrgyz (written \cyrill{``кыргыз тили'' {or ``кыргызча''},} and alternatively as ``Kirghiz'' or ``Kirgiz'') is a Turkic language spoken in Kyrgyzstan, China, Tajikistan, Uzbekistan, and several other countries. Its classification within Turkic is complex; it appears to belong alternately to the Kypchak (Northwestern) branch and the South Siberian (Northeastern) branch. Phonetically and phonologically, the Turkic varieties most similar to Kyrgyz are the southern dialects of Altay, though Kyrgyz shows strong parallels to Kazakh, especially in its Talas (Talas Region) dialects, which these Altay varieties lack~\cite{washington2012finite}. In Southern varieties of Kyrgyz, there are also notable similarities to Uzbek that other dialects lack.

Kyrgyz is spoken primarily in Kyrgyzstan, where it holds official status as the national language. Many residents of Kyrgyzstan are bilingual in Kyrgyz and Russian and/or Uzbek, comprising a majority of the country's population. In Central Asia, sizable Kyrgyz-speaking communities exist outside Kyrgyzstan, most notably in China (where the Kyrgyz are an officially recognized minority), Tajikistan, Afghanistan, Pakistan, and Uzbekistan. The current estimate of the number of speakers is approximately 5.3 million,\footnote{Based on figures from~\cite{nationalstatcom}.} primarily in Kyrgyzstan.

The Kyrgyz language was initially written in the Göktürk script, which was later replaced by the Perso-Arabic alphabet. This script was used until 1928 in the Soviet Union and is still in use in China, Afghanistan, and Pakistan. From 1928 to 1940, the Uniform Turkic Alphabet, based on the Latin script, was introduced. However, in 1940, Soviet authorities mandated the use of the Cyrillic alphabet for all Turkic languages within their territory. After Kyrgyzstan gained independence in 1991, there was renewed interest in transitioning to the Latin alphabet. Although this shift has not yet been implemented, it remains a topic of occasional discussion. Examples of Kyrgyz writing in different scripts, including historical ones, are presented in Table~\ref{tab:scripts}.

\begin{table}[H]
    \centering
    \begin{tabular}{|p{2.4cm}|p{2.5cm}|p{2.2cm}|p{2.5cm}|p{2.2cm}|}
    \hline
    \centering\textbf{Cyrillic alphabet (1938 -- now)} & 
    \centering\textbf{Old Latin alphabet (1928 -- 1938) } &
    \centering\textbf{Arabic alphabet (pre 1928) } & 
    \centering\textbf{Old Turkic alphabet (VIII -- X c.) } & 
     \centering \textbf{Translation into English language} \tabularnewline
    \hline    
        \raggedright\begin{spacing}{0.6}\cyrill{\scriptsize Бардык адамдар өз беделинде жана укуктарында эркин жана тең укуктуу болуп жаралат. Алардын аң-сезими менен абийири бар жана бири-бирине бир туугандык мамиле кылууга тийиш.}\end{spacing} & 
        \raisebox{-\totalheight}{\includegraphics[width=1\linewidth]{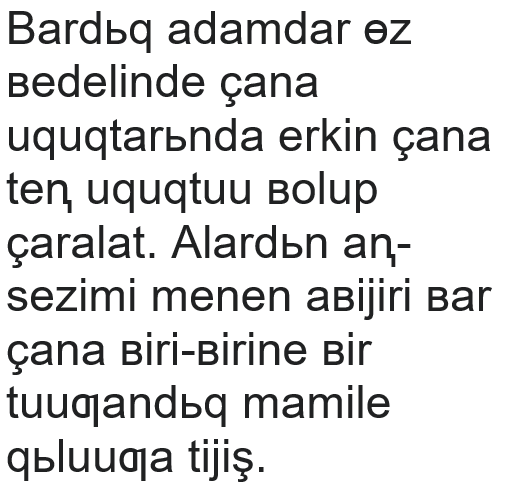}} & 
        \raisebox{-\totalheight}{\includegraphics[width=1\linewidth]{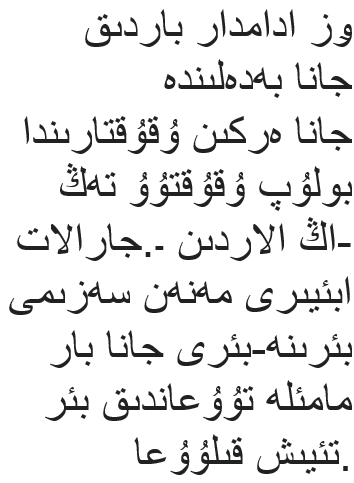}} 
        & \raisebox{-\totalheight}{\includegraphics[width=1\linewidth]{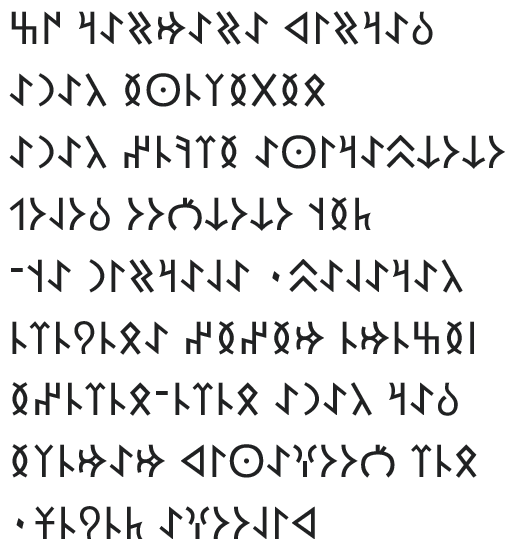}} & 
        \raggedright\begin{spacing}{0.6}{\scriptsize  All human beings are born free and equal in dignity and rights. They are endowed with reason and conscience and should act towards one another in a spirit of brotherhood.}\end{spacing}  \tabularnewline\hline
    \end{tabular}
    \caption{The examples of Kyrgyz scripts~\cite{wikipedia:kyrgyz_language}.}
    \label{tab:scripts}
\end{table}

\paragraph{Certain Linguistic Characteristics.}
Kyrgyz uses a subject-object-verb word order and lacks grammatical gender, with gender conveyed through context. Unlike English, Kyrgyz does not have several analytical grammatical features, such as auxiliary verbs (e.g., ``to have''), definite and indefinite articles (e.g., ``the'' and ``a/an''), modal verbs (e.g., ``should'', ``will''), dependent clauses, and subordinating conjunctions (e.g., ``that'', ``before'', ``while''). Instead, it replaces these with various synthetic grammatical structures.

Kyrgyz is notably agglutinative, meaning it forms words by stringing together morphemes without altering their spelling or phonetics. This structure results in a vast number of possible word forms. While a verb in Russian can have a maximum of 150 written forms, in Kyrgyz, this number can reach thousands. A famous illustrative example: one of the legitimate interrogative forms of the verb ``\cyrill{\textbf{камсызда}}'' (``to provide,'' ``to guarantee,'' ``to insure'') is ``\cyrill{\textbf{камсыздандырылбагандардансыздарбы}}'' (``Are you from those who have not been insured?'').

Prominent phonological features, such as vowel harmony and consonant assimilation, also require special handling in rule-based NLP methods.

\subsection{Linguistic Relations}

As noted above, classification of the Kyrgyz language within the Turkic family remains problematic. The Swadesh 100-word list~\cite{swadesh1952lexicostatistic} is often used to determine the degree of lexical similarities and differences between languages as a percentage. This list includes words from the basic lexicon, which is generally resistant to historical changes across languages worldwide. A higher percentage of matching words between languages indicates closer linguistic kinship.

Results of lexical-statistical analysis of the Kyrgyz language's relationship to certain Turkic languages, based on the 215-word Swadesh list~\cite{swadesh1955accuracy,lindsay2016mutual}:
\begin{inparaenum}[(1)]
    \item~\textbf{Kazakh}: 91\%,
    \item \textbf{Tatar}: 79\%,
    \item \textbf{Uyghur}: 77\%,
    \item \textbf{Uzbek}: 76\%,
    \item \textbf{Altai}: 73\%.
\end{inparaenum}

A lexical similarity of over 85\% suggests that two languages in comparison are likely related as dialects of a single language. These similarities indicate significant potential for transfer learning and resource sharing among Turkic languages.


\section{Challenges in Kyrgyz NLP}\label{sec:challenges}

\paragraph{Scarcity of Machine-Readable Resources.}
A core challenge for any LRL is the shortage of large, annotated text corpora for training and evaluating models, as well as other structured resources such as dictionaries, lexicons, and thesauri. For Kyrgyz, an additional issue arises from the limited diversity in easily obtainable data (e.g., from web scraping), which is predominantly sourced from news articles. This focus on news media restricts the range of language styles and registers available for analysis and model training.

\paragraph{Multiple Scripts and Dialects.}
As discussed in Section~\ref{ssec:kyrgyz101}, Kyrgyz speakers use multiple scripts. Ideally, tools should support both Cyrillic and Perso-Arabic scripts directly or incorporate reliable open-source transliteration solutions~\cite{washington2020multi} within user-friendly interfaces.\footnote{A suitable tool is already available. A user-friendly interface is implemented for Kazakh in \emph{apertium-kaz}; a similar adaptation for \emph{apertium-kir} is considered feasible by the authors: \url{https://github.com/apertium/apertium-kir/issues/13}} Additionally, dialectal variations affect vocabulary, pronunciation, and grammar. For instance, the ``Northern vs Southern'' dialectal division in Kyrgyz shows the influence of Mongolian and Uzbek, respectively.

\paragraph{Agglutinative Morphology.}
As highlighted in Section~\ref{ssec:kyrgyz101}, Kyrgyz’s agglutinative nature generates a vast number of word forms. A single verb in Kyrgyz, for example, can have thousands of forms, compared to a maximum of around 150 in Russian~\cite{tyers_morphological_modelling,aytnatova_kyrgyz_language_2016}. This complexity requires sophisticated morphological analysis algorithms to parse and understand word structures.

\paragraph{Decentralized Initiatives.}
Previous corpus-building efforts have been fragmented (for details, see Section~\ref{ssec:cor}), resulting in duplicated work and a lack of standardization across initiatives. Developing a comprehensive national corpus remains an ongoing effort.

Despite these challenges, there is a growing demand for NLP applications in Kyrgyz. With over 5 million native speakers and official status in Kyrgyzstan, there is a need for tools such as machine translation, speech recognition, and text analysis. The increasing digital presence of Kyrgyz speakers further underscores the need for advancements in NLP to support communication, education, and access to information.

\section{Existing Surveys of the Field}\label{sec:existing_surveys}

Research on computational approaches to processing the Kyrgyz language spans several decades. To the best of our knowledge, only two surveys focused on computational linguistics aspects of Kyrgyz text processing have been published prior to this work.

In the study by Musayev~\cite{problems2013musayev} presented at the first \emph{TurkLang} conference, the state of computational linguistics in Kyrgyzstan (as of 2013) is reviewed, covering developments since the 1990s. According to the authors, the key areas requiring development are
    \begin{inparaenum}[(1)]
        \item unified word formation algorithms,
        \item creation of computational language models and linguistic resources,
        \item standardization of orthography and spelling,
        \item transition to the Latin alphabet for Turkic languages.
    \end{inparaenum}
The survey also identifies key interest groups and institutions involved in Kyrgyz NLP, such as the Institute of Theoretical and Applied Mathematics of the National Academy of Sciences of Kyrgyzstan and the computational linguistics department at the Kyrgyz State University of Construction, Transportation, and Architecture (as of 2024, part of KSTU named after I. Razzakov). TamgaSoft's\footnote{\url{https://tamgasoft.kg/en/}} solutions are mentioned.

In a recent 2024 survey of Central Asian Turkic languages by Y. Veitsman~\cite{veitsman2024recentadvancementschallengesturkic}, progress in data collection and NLP for Kyrgyz is acknowledged, though significant challenges remain. The author recommends collecting more high-quality data, focusing on transfer learning from related languages (e.g., Kazakh and Turkish), and leveraging LLMs for data augmentation. Veitsman’s survey categorizes Kyrgyz as ``Scraping By'', following the classification in~\cite{joshi2020state}, a label previously assigned in~\cite{mirzakhalov2021large} as well. This designation indicates insufficient resources and initiatives for Kyrgyz NLP, attributed to factors such as a lack of language-focused initiatives, the dominance of Russian as a \textit{lingua franca} in Central Asia, and limited internet access in some regions.

\section{Scientometric Analysis of Kyrgyz NLP}\label{sec:scientometrics}

To elucidate the landscape of Kyrgyz NLP research, we conducted a scientometric analysis of $67$ research works. While some studies date back to the 1980s and 1990s,\footnote{The most notable contributions from this period include the seminal works of~\textbf{Dr. Tashpolot Sadykov}, such as~\cite{sadykov1987problemy,sadykov1995foundations}. A distinguished Turkologist and computational linguist, Dr. Sadykov passed away in November 2023 and would have turned 75 in October 2024. His pioneering contributions remain foundational to the field, and his absence is profoundly felt by the research community.} our primary focus has been on works published from 2011 onward. This choice is due to the limited availability of programs or datasets from earlier publications, which restricts their applicability to contemporary computational practices.

Our data collection methodology proceeded as follows.

We initially gathered papers containing the keyword ``Kyrgyz'' in their title or abstract from all \emph{TurkLang} conferences\footnote{\emph{TurkLang} is an annual conference dedicated to research on Turkic languages, typically held in regions with a significant population of Turkic language speakers.}, along with all works cited within these papers and publications from the authors’ portfolios. After excluding irrelevant studies (e. g., non-NLP papers or those lacking computational methods), additional papers were identified through~\emph{Google Scholar} and other academic databases.\footnote{Also note that, in 2024, the ACL SIGTURK workshop has been established: \url{https://sigturk.github.io/workshop/202400/}}

A primary limitation of this approach is that it does not ensure comprehensive coverage of all Kyrgyz NLP research. However, we believe this compilation captures the contributions of researchers actively engaged in the field and invested in the visibility and impact of their work. Our collection encompasses papers presented at relevant events or in relevant journals and works clearly indicating connection to the Kyrgyz language, readily accessible through permissive indexing engines (contrasting with more restrictive databases such as Scopus or Web of Science).

Another notable limitation to consider when interpreting the following analysis is the exclusive focus on Kyrgyz NLP, rather than Turkic NLP (including multilingual projects leveraging commonalities among Turkic languages, such as~\cite{gatiatullin2020turkic}). A broader analysis would have required substantially more time and likely necessitated a more extensive study beyond the scope of this paper, which was derived from a keynote presentation constrained to 40 minutes.

\subsection{Categorization of Topics.} The collected works were manually categorized by the authors into the following branches.

\begin{itemize}
    \item[\texttt{MOR}] Studies focused on morphological analysis and modeling. 
    \item[\texttt{SYN}] Research addressing syntactic structures and parsing. 
    \item[\texttt{SEM}] Works related to semantic analysis and word meaning. 
    \item[\texttt{COR}] Efforts in building and analyzing text corpora. 
    \item[\texttt{DIC}] Construction and usage of lexical resources. 
    \item[\texttt{APP}] Applications, such as machine translation and topic classification. 
    \item[\texttt{MSC}] General surveys and other relevant studies. 
\end{itemize}
During annotation, the~\texttt{APP} tag was assigned only when none of the other labels above (\texttt{MOR}, \texttt{SYN}, \texttt{SEM}, \texttt{COR}, \texttt{DIC}) applied to the paper. If none of the categories except \texttt{MSC} were suitable, the~\texttt{MSC} label was used. Some manuscripts were assigned multiple labels to reflect their interdisciplinary nature.

\begin{figure}[H]
    \centering
    \includegraphics[width=0.99\linewidth]{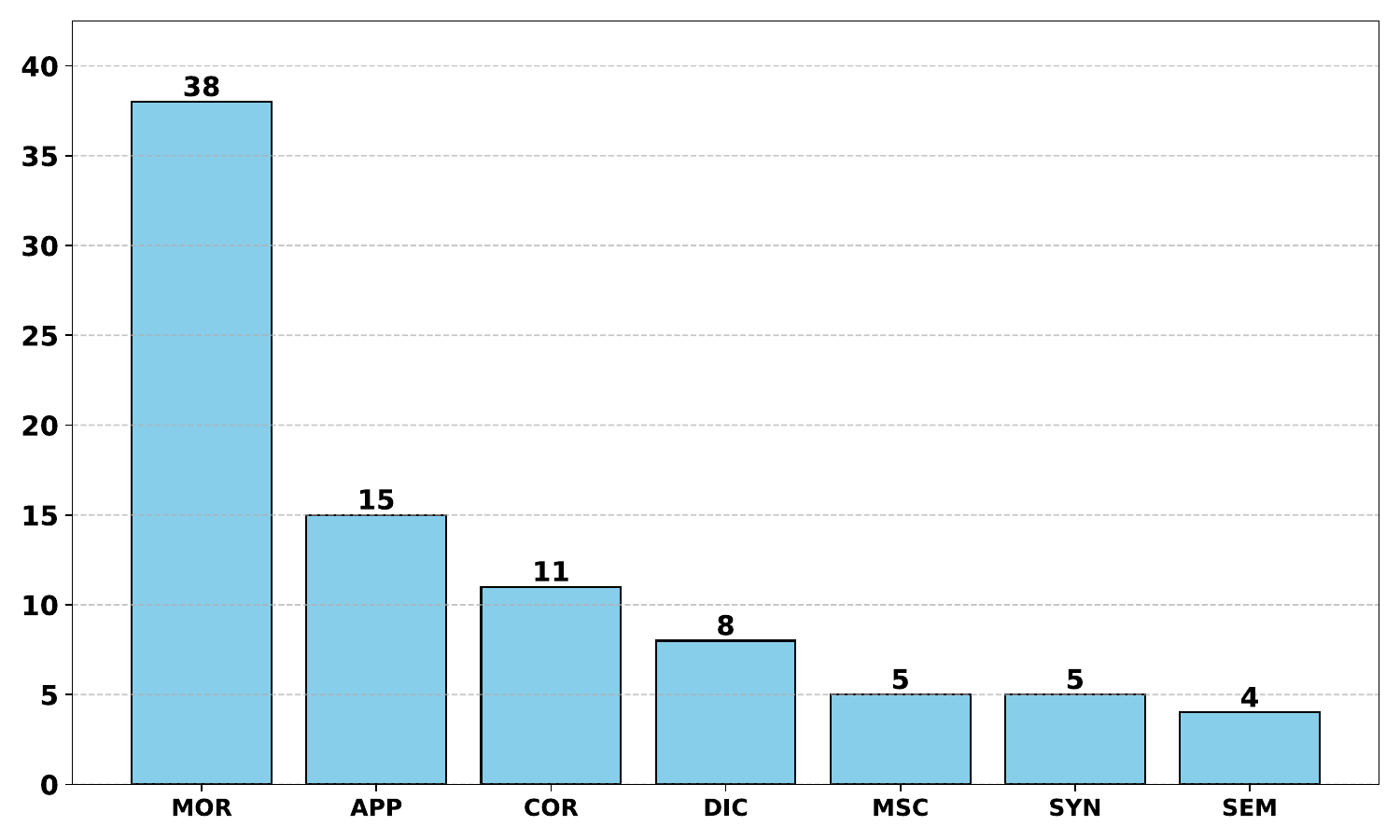}
    \caption{Distribution of the total number of labels (note: some articles may have multiple labels, so the total does not equal the number of articles).}
    \label{fig:histogram_topics}
\end{figure}

As expected, works on morphology are the most prevalent, beginning with the earliest paper in our dataset, while Treebank-related studies have only recently begun to constitute a notable proportion. Although there are relatively few papers on semantics, they exhibit considerable diversity in both approach and focus. Several books and surveys, despite their broad titles, do not claim to be comprehensive (see the statistics in Fig.~\ref{fig:histogram_topics}).

\paragraph{Collaboration Networks.} The manual analysis of co-authorship revealed the following collaboration patterns:
\begin{inparaenum}[(1)]
    \item \textbf{Institutional Collaborations}: researchers from the same institution often co-authored papers;
    \item \textbf{International Cooperation}: some projects involved international teams, particularly in areas like machine translation and universal dependencies;
    \item \textbf{Interdisciplinary Teams}: successful projects often combined expertise in linguistics, computer science, and data science.
\end{inparaenum}

\begin{figure}[H]
    \begingroup
    \selectlanguage{english}
    \fontsize{6}{9}\selectfont 
    \centering

	\includegraphics[width=1\linewidth]{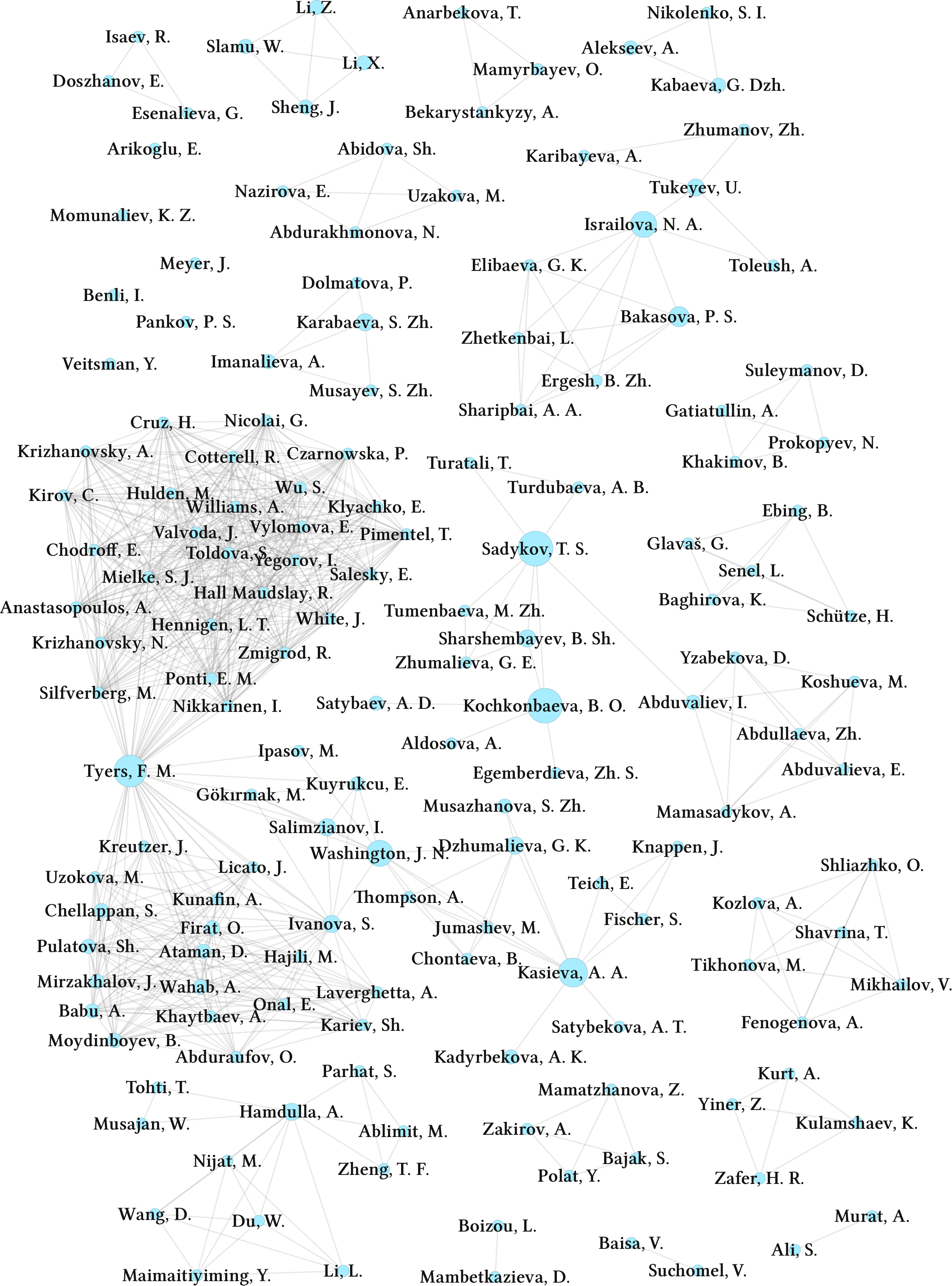}
    \caption{Collaboration (co-authorship) network based on the Kyrgyz NLP bibliography. Nodes represent individual authors, and an edge between two nodes indicates co-authorship on at least one work.}
    \label{fig:coauthorship}
    \endgroup
\end{figure}

\subsection{Observations}

Most cliques in the co-authorship network represent papers with multiple authors. Isolated vertices indicate single-author papers, i. e., works by researchers who contributed to Kyrgyz NLP only once (in our dataset), typically with either a survey or a focused, standalone contribution. Connected components that are not cliques (i.e., involving multiple papers) offer insights into collaboration patterns.

Unsurprisingly, the nodes in this study tend to cluster around specific tasks or institutions, reflecting researchers’ primary affiliations or project focuses.

In examining the relationship between collaboration structure, topics, and types of studies presented in the paper, we observed the following patterns.

\textbf{Resource Scarcity.} There is a high demand for tools and datasets, and contributions in these areas are well-received in the research community.
 
\textbf{Online Presence.} Research visibility is affected by online presence; projects without digital traces were less likely to be included or cited.

\textbf{Complex Studies Require Large and Diverse Research Teams.} The largest communities in the graph, such as those around the SIGMORPHON paper~\cite{vylomova2020sigmorphon} and Turkic Interlingua studies~\cite{mirzakhalov2021evaluating,mirzakhalov2021large}, exemplify the need for large, collaborative efforts. While certain individuals may drive these projects, their success depends on the collective work of interdisciplinary teams.

\textbf{Textbooks are Frequently Cited.} Although we did not conduct a detailed citation analysis, textbooks and educational materials are frequently cited, suggesting their value to the field. The development of relevant textbooks, courses, and workshops is in high demand and could facilitate the adoption of modern NLP methods, accelerating field progress.

Since we have published the relevant data online\footnote{Bibliography in JSON format: \url{https://kyrgyznlp.github.io/static/bibliography_joined.json}.}, we encourage those interested in formal collaboration analysis to apply Social Network Analysis (SNA) methods~\cite{wasserman1994social,scott2017sna} to the collaboration graph, labeled bimodal author-paper graph, citation graph (not yet collected), etc. Although the graph sizes are small in this case, comparing indices such as betweenness centrality with other indicators of researchers' output impact could provide a more conclusive basis to support or challenge our observations.


\section{Overview of Kyrgyz NLP Research}\label{sec:overview_kyrgyznlp}

\subsection{Morphology and Subword-Aware Processing Studies (MOR)}\label{ssec:mor}

\paragraph{Finite-State Transducers (FSTs)} have been employed to model Kyrgyz morphology due to their suitability for agglutinative languages. FSTs efficiently handle the complex affixation and morphological variations. An effective and open morphological analyzer was introduced in 2012 by J. N. Washington, F. M. Tyers, and M. Ipasov~\cite{washington2012finite} as part of the Apertium project~\cite{forcada-tyers-2016-apertium}. This tool, along with other updates and relevant works~\cite{washington2014finite,washington2019free,washington2020multi}, has been made publicly available as code on GitHub.\footnote{E. g.,~\url{https://github.com/apertium/apertium-kir}}

\paragraph{Rule-Based Modeling of Different Aspects of Morphology.}
Researchers have developed rule-based systems to capture morphological rules since at least the 1980s–1990s~\cite{sadykov1987problemy,pankov1992obuch}. These systems define patterns for stem and affix combinations, addressing issues such as vowel harmony and consonant assimilation. Various aspects of morphological modeling along these lines have been explored since then~\cite{sadykov2017model,kochkonbaeva2018morpho,sadykov2018morfologicheskie,sadykov2018optim,israilova2019,satybaev2019testirovanie,satybaev2019math,kochkonvaeva2020modeling,nazirova2023morph}, though only a few reference open implementations or datasets~\cite{sadykov2023morph}, or are integrated into larger frameworks~\cite{yiner2016kyrgyz}. Also notable are foundational books on the subject~\cite{sadykov1995foundations,abduvaliev1997azyrky,sadykovbook2015}.

Other algorithmic approaches to morphology modeling involve complex data structures, ontology-like knowledge bases, and various tasks, such as updating vocabulary in information systems~\cite{bakasova2016,israilova2017morphological,israilova2017algo,sharipbay2018comparison}.

Another recently studied direction in morphology modeling is the CSE (Complete Set of Endings) approach, discussed in works~\cite{tukeyev2020,toleush2021development,tukeyev2023new}. In particular,~\cite{tukeyev2020} explores its applications for neural machine translation.

\paragraph{Perso-Arabic Script} as a medium for the Kyrgyz language (primarily used in China) has also attracted researchers' attention. Publicly available studies are few, but they include works on agglutination-aware model fine-tuning~\cite{li2020agglutifit,li2020low} and phonological/morphological tools for a uniform multilingual information processing platform primarily targeting \textit{UKK (Uyghur, Kazakh, Kyrgyz) languages}~\cite{ablimit2017multilingual}.

Works on character-level approaches to morphology are limited; the only example we have found is~\cite{ali2023improved}.

\subsection{Syntax (SYN)}\label{ssec:syn}

Several recent projects have adapted the~\emph{Universal Dependencies} framework~\cite{ud2021} for Kyrgyz. For links and descriptions, please see Section~\ref{sec:available_resources}.

Additional efforts are underway to build syntactically annotated corpora and establish rules and procedures for annotators (e.g., handling copula tokenization, small words, null-headed clauses, and distinguishing inflection from derivation). This task is as complex as similar projects for other Turkic languages and has been addressed in several recent works~\cite{dzhumalieva2023,musazhanova2023,kasieva2023problems}. These resources are essential for training parsers and developing syntactic analysis tools.

\subsection{Semantics (SEM)}\label{ssec:sem}

Studies focused on semantics in Kyrgyz are few and diverse in tasks and approaches. Works~\cite{karabaeva2015computer,karabaeva2016virt} describe computational methods for directly modeling spatial semantics in Kyrgyz, focusing on spatial terms, grammar, and the virtual geometrical spaces created by verbs. The paper~\cite{kasieva2022disambiguation} is the only known attempt at word sense disambiguation (WSD) for Kyrgyz, particularly addressing verb sense disambiguation (VSD). Rule-based methods for VSD automation are presented, which hold value for NLP and corpus linguistics in agglutinative languages such as Kyrgyz. A recent effort to create a word embeddings evaluation dataset via direct translation of the RUSSE benchmark has also been published~\cite{alekseev2023embeddings}, with several models evaluated. An even more recent work~\cite{shliazhko2024mgpt} on multilingual GPT models was accompanied by the release of the \texttt{mGPT-1.3B-kirgiz} model.\footnote{The model is freely available online: \url{https://huggingface.co/ai-forever/mGPT-1.3B-kirgiz}}

\subsection{Corpus Studies (COR)}\label{ssec:cor}

Research mentioning corpora is also limited, with some studies noted above (regarding morphological and syntactic annotation for Kyrgyz corpora).

The study~\cite{sadykov2013manas} discusses creating a national corpus of the Kyrgyz ``Manas'' epic, proposing corpus linguistics techniques to analyze and document its vocabulary and grammar. A framework for building a searchable dictionary of the epic's texts, useful for linguistic research and preserving Kyrgyz heritage, is described, though no code is provided, and the corpus is still under development.

In the paper~\cite{baisa2015turkic}, Turkic language support in Sketch Engine is described, focusing on building corpora and providing tools for analysis in Kazakh, Kyrgyz, and Turkish. The paper presents methods for web-crawling, concordance searches, and word sketches.

Papers~\cite{kasieva2020pos,kasieva2020posapertium,kasieva2021corpus} focus on part-of-speech (POS) annotation for a newly created Kyrgyz language corpus, utilizing the Turkic Lexicon Apertium platform, with detailed examples of tagging and morphological analysis.

\subsection{Applied NLP (APP)}\label{ssec:app}

The works focusing on applied tasks (mostly of direct industrial value) are, as expected, diverse. While some studies concentrate on rule-based approaches to aid application development, others employ recent neural approaches.

\paragraph{Analysis and Rule-Based.} The work~\cite{polat2018machine} compares the accuracy of Google Translate and Yandex Translate for translating Kyrgyz proverbs into English and Turkish, focusing on lexical, semantic, and syntactic performance. It concludes that Google Translate performs better overall, particularly in handling Kyrgyz's agglutinative structure, though both systems struggle with proverbs. No code or data are provided, but the presented error analysis highlights key challenges for improving Kyrgyz machine translation. In the works~\cite{kochkonbaeva2016,buazhar2018automatic}, an algorithm for machine translation from Russian to Kyrgyz, among others, is discussed; however, these texts primarily focus on morphological and syntactic analysis. They provide a framework for processing affixes and word formation in Kyrgyz, which may improve translation accuracy between these languages in automated systems. These papers mention that a morphological analyzer was developed using Delphi with a database of common Russian words, although no specific code is provided.

A 2019 overview~\cite{washington2019free} describes the free and open-source technologies for Turkic languages developed in the Apertium project~\cite{forcada-tyers-2016-apertium}, including morphological transducers and machine translation systems for languages such as Kazakh, Kyrgyz, and Tatar. These tools support language revitalization and linguistic rights and are useful for developing NLP resources in under-resourced Turkic languages. All code is available, and the technologies are production-ready for several languages.

\paragraph{Neural Machine Translation.}

Two notable works~\cite{mirzakhalov2021evaluating,mirzakhalov2021large} evaluate multiway multilingual neural machine translation (MNMT) for Turkic languages, covering 22 languages, most of them less-resourced. The comparison between bilingual baselines and MNMT models shows that MNMT performs better in out-of-domain tasks, which is useful for enhancing machine translation in under-resourced Turkic languages. Interestingly, the results also indicate that bilingual baselines often outperform MNMT models for Kyrgyz in domain-specific translations. All code and models are publicly available.

\paragraph{Other Applications}

The work~\cite{tohti2008character} proposes solutions for character code conversion and spelling error correction in Uyghur, Kazakh, and Kyrgyz, aimed at multilingual information retrieval. Methods for converting non-Unicode characters to Unicode and root-based query expansion to handle spelling errors are introduced, enhancing search accuracy and recall in these languages. Although no specific code is provided, it is claimed that the algorithms for character conversion and query correction have been tested within the system.

A more recent study~\cite{alekseev2023benchmarking} presents a manually labeled topic multi-label classification dataset of news articles collected from the news site \textit{24.kg}. Several baseline models are evaluated, with results showing that multilingual neural models like \texttt{XLM-RoBERTa} perform best, which is valuable for advancing Kyrgyz-language NLP tools. The dataset and models are planned for public release. Additionally, the construction of a stopwords list (stoplist) — words considered insignificant for various NLP tasks — is discussed in~\cite{isaev2023towards}.

A notable contribution, the benchmark described in~\cite{senel2024kardecs} focuses on cross-lingual transfer from Turkish to less-resourced Turkic languages like Kyrgyz. The work presents strategies for improving natural language understanding tasks through intermediate training and fine-tuning with Turkish, showing significant accuracy gains for Kyrgyz, which holds potential for future NLP advancements. All code and models are publicly available for further research and applications.


\section{Existing Resources}\label{sec:available_resources}

\paragraph{Dictionaries}  for Kyrgyz are numerous, listing all of them is beyond the scope of this paper, however, we would like to emphasize that making them machine-readable is essential for further progress in corpus creation and overall Kyrgyz NLP.

\paragraph{Corpora.}
\textit{Manas-UdS} corpus~\cite{kasieva2020poster} contains $1.2$ million words from $84$ literary texts across five genres. It includes lemmatization, part-of-speech tags, and rich metadata. \textit{kyWaC}, a web corpus of Kyrgyz comprising $19$ million words collected in January 2012~\cite{baisa2015turkic}. \textit{Leipzig Corpora Collection}~\cite{kir_newscrawl_2011,kir_community_2017,kir-kg_web_2019,kir_news_2020,kir_wikipedia_2021} offers community data, news crawl data, and Wikipedia texts totaling hundreds of thousands of sentences. \textit{TilCorpusu}: a 300 million-word corpus including news and fiction data made public in December 2023\footnote{As of now, already available at: \url{https://huggingface.co/datasets/the-cramer-project/Kyrgyz_News_Corpus}.}.

\paragraph{Syntax Treebanks.}
\textit{UD\_Kyrgyz-KTMU}: a treebank with $781$ sentences annotated with dependency relations~\cite{benli2023}; the description has not been yet provided, though it is clear that the unspecified sources are various: translations of E. Hemingway's works to Kyrgyz, news articles, etc. the percentages of each genre are yet to be determined.
\textit{Kyrgyz-TueCL} treebank contains a total of $145$ sentences (including $20$ \textit{Cairo} sentences~\cite{nivre2015towards}), and $\approx$ $100$ sentences suggested by the~\emph{UD Turkic Group}~\footnote{For more information, please see \url{https://github.com/ud-turkic}.}. Translations of all sentences are provided in English, Turkish and Azerbaijani languages; this dataset is a part of the UD Turkic Treebank. 

\paragraph{Other Resources} are focused on narrow tasks. \textit{WikiANN} dataset contains a Kyrgyz subset for NER tasks, and~\textit{KyrgyzNER} is a dataset currently under development, expected to enhance NER capabilities. Used for training respectively \texttt{murat/kyrgyz\_language\_NER}\footnote{\url{https://huggingface.co/murat/kyrgyz_language_NER}} and~\texttt{TTimur/\allowbreak xlm-roberta-base-kyrgyzNER}\footnote{\url{https://huggingface.co/TTimur/xlm-roberta-base-kyrgyzNER}}\footnote{\textit{KyrgyzNER} project heavily depended on the volunteers' and students' efforts and came into being thanks to the collaboration with G. Kabaeva and G. Zhumalieva (KSTU n. a. I. Razzakov).}

The previously mentioned word similarity dataset, \textit{HJ-Ky-0.1}, for evaluating word embeddings~\cite{alekseev2023embeddings}, is the first benchmark of its kind for Kyrgyz. It is yet to be published following the competition event, along with the first topic classification benchmark~\cite{alekseev2023benchmarking}.

Verbal paradigms for Kyrgyz (100 Kyrgyz verbs fully conjugated in all tenses)~\cite{aytnatova_kyrgyz_language_2016} were prepared and published\footnote{\url{https://github.com/unimorph/kir}} for Unimorph by E. Chodroff~\cite{vylomova2020sigmorphon}.

For LLM-related evaluation and training, Stanford Alpaca~\cite{taori2023alpaca} instructions translated into Kyrgyz using ChatGPT and Google Translate are available online.\footnote{\url{https://github.com/Akyl-AI/kyrgyz-alpaca}}

The model \texttt{mGPT-1.3B-Kirgiz}~\cite{shliazhko2024mgpt}, previously mentioned in Section~\ref{ssec:sem}, has also been made public.

A joint project by Ulutsoft LLC and the National Commission on the State Language and Language Policy has resulted in the release of several models on HuggingFace: a text completion model \texttt{Mistral-7B-v0.1-kyrgyz-text-completion}\footnote{\url{https://huggingface.co/UlutSoftLLC/Mistral-7B-v0.1-kyrgyz-text-completion}}, a text-to-speech model \texttt{kyrgyz-tts}\footnote{\url{https://huggingface.co/UlutSoftLLC/kyrgyz-tts}}, and a speech recognition model \texttt{whisper-small-kyrgyz}\footnote{\url{https://huggingface.co/UlutSoftLLC/whisper-small-kyrgyz}}. While speech processing is not often classified as NLP, related techniques can play a crucial role in data collection and verification for less-resourced languages.

The Cramer Project's \textit{AkylAI}, a smart voice assistant made possible by volunteer efforts, is accompanied by an open text-to-speech model\footnote{\url{https://huggingface.co/the-cramer-project/TTS_small}} and several other models relevant to this study, available at~\href{https://huggingface.co/the-cramer-project/}{huggingface.co/the-cramer-project} and~\href{https://github.com/Akyl-AI}{github.com/Akyl-AI}.

For more resources, one may consult an ``awesome list'' of references: \url{https://github.com/alexeyev/awesome-kyrgyz-nlp}; this registry is regularly updated, and anyone is welcome to contribute via a pull request.


\section{Untapped Subdomains of KyrgyzNLP}\label{sec:untapped}

Having described the corpus of publicly available research outcomes, it makes little sense to list the unexplored areas, as ``everything else'' remains largely uncharted. However, we aim to highlight the most conspicuous gaps.

\subsection{Lacking or Unavailable Language Resources}

The variety of language resources required is vast. Here, we aim to emphasize the necessity of specific datasets and other helpful resources, the lack of which obstructs multiple paths for Kyrgyz NLP research.

\begin{itemize}
    \item[\textbf{Dictionaries}] are basic yet essential language resources. Although work on new dictionaries is ongoing~\cite{arycoglu2021} and many human-readable dictionaries exist~\cite{koshueva2021kyrgyz} (including OCR-ready PDFs), converting them to machine-readable formats is crucial. A recent dictionary conversion~\cite{kasieva2024structured} and earlier works~\cite{momunaliev2016,boizou2017kyrgyz} offer hope that such efforts will yield resources more easily than before. However, the core issue is licensing, and obtaining permissions for other dictionaries and lexicons remains essential for making these tools truly open and usable.

    \item[\textbf{Transliteration}] tools for languages with multiple scripts, influenced by region or other factors, are partly available~\cite{washington2020multi}, as highlighted in Section~\ref{sec:challenges}. More user-friendly interfaces, benchmarks, and publicly available datasets in scripts other than Cyrillic are still needed.

    \item[\textbf{Corpora},] including a comprehensive National Corpus, are crucial for language analysis, historical linguistics, and establishing language norms. We discuss corpus collection and annotation further in Section~\ref{sec:roadmap}. Additionally, larger \textbf{parallel corpora} are needed to improve machine translation to and from Kyrgyz, as indicated by the topics covered in Section~\ref{sec:overview_kyrgyznlp}. From our experience, many Kyrgyz-language-aware machine translation tools lack established glossaries, leading to frequent borrowing of terms from Russian or English. While there have been efforts in~\textbf{terminology development}~\cite{zhumalieva2018lingvisticheskii} and~\textbf{domain-specific dictionaries}, e.g.,~\cite{law2014turar}, significant work remains in this area.

    \item[\textbf{Information Extraction}] datasets, including those for nested NER, OpenIE, relation extraction, temporal information extraction, entity linking, coreference resolution, etc., are crucial for training and evaluating models that can support both academic research (e.g., corpus annotation) and industry applications.

    \item[\textbf{Task-Specific QA \& Reading Comprehension}] datasets, along with other tasks often considered ``AI-hard,'' are in high demand. As discussed in Section~\ref{sec:data_nlp}, Large Language Models for Kyrgyz are on the horizon, and these models require evaluation with task-specific data.
\end{itemize}

Beyond news and encyclopedic content, \textbf{legal documents data} for Kyrgyz are increasingly available online. Legal NLP is a growing field (see \href{https://github.com/maastrichtlawtech/awesome-legal-nlp}{awesome-legal-nlp}, \url{https://nllpw.org/}, and many related surveys on Google Scholar). The need for intelligent processing of official documentation will likely grow in Kyrgyzstan as well, making this a potentially important area to prioritize in the near future.

\subsection{Untried Methods}

From the methodological point of view, as highlighted in Sections~\ref{sec:data_nlp} and~\ref{sec:existing_surveys}, a large number of techniques traditionally applied for less-resourced languages has not been implemented or employed. These techniques could significantly speed up the progress in the field: transfer learning (beyond multiway translation), cross-lingual models in general, data augmentation and synthesis (beyond the automatic translation, which is already used), etc.

We list some of the tasks that can aid in corpora annotation and provide the comments how exactly they can be of some use for the purpose.

\begin{itemize}
    \item[\textbf{Named entity recognition (NER):}] automatically labeling entities like names, locations, dates, and organizations in text, NER could provide baseline annotations for entities, which annotators can then validate or adjust.

    \item[\textbf{Part-of-speech tagging:}] assigning PoS tags to words in text is a fundamental task that serves as a foundation for more complex syntactic and semantic annotations.

    \item[\textbf{Dependency parsing:}] this task analyzes the grammatical structure of sentences, assigning tree structures that can be used to automate or guide syntactic annotation processes; the works on treebanks are in progress, please see the information about the current state in Sections~\ref{ssec:syn} and~\ref{sec:available_resources}.
    
    \item[\textbf{Coreference resolution:}] identifying when different expressions refer to the same entity (e. g., ``Meerim's'' and ``her'' in the same context) aids in creating coreference annotations by automating entity linking and anaphora resolution.

    \item[\textbf{Semantic role labeling (SRL):}] semantic roles are assigned to parts of a sentence, like identifying agents, themes, and goals (e. g., ``who did what to whom''), which assists in annotating event and argument structures.

    \item[\textbf{Topic modeling:}] unsupervised methods like Latent Dirichlet Allocation (LDA)~\cite{blei2003latent}, BERTopic~\cite{grootendorst2022bertopic}, etc. can group texts by topic, helping annotators quickly identify and label topic-specific text segments within large corpora.

    \item[\textbf{Speech segmentation and speaker diarization:}] segmenting audio into units like words, phrases, and sentences, and distinguishing between speakers, facilitates annotation of multi-speaker conversations or interviews, which is essential e. g. in building the spoken part of the corpus.

    \item[\textbf{Word sense disambiguation (WSD):}] identifying the correct sense of a polysemous word based on context aids in annotating corpora where accurate word senses are crucial for semantics.

    \item[\textbf{Event detection and temporal tagging:}] identifying events, their participants, and temporal information supports the annotation of chronological narratives and event-based corpora, often used in news and historical text.

    \item[\textbf{Text classification for domain labeling:}] automatically assigning documents to predefined categories (e.g., news, sports, science) helps streamline the labeling process in domain-specific corpora.

    \item[\textbf{Paraphrase detection:}] identifying similar phrases or sentences can assist annotators by flagging repetitive content, reducing manual work when labeling similar or redundant text.

    \item[\textbf{Automatic summarization:}] creating summaries for large documents can help annotators quickly grasp content, especially for annotation tasks that involve long texts.

    \item[\textbf{Dependency and alignment mapping for parallel texts:}] for bilingual or multilingual corpora, aligning segments between two languages can aid annotation, especially for translation or language comparison tasks.

    \item[\textbf{Active learning and uncertainty sampling:}] ML models can suggest samples that are uncertain or difficult for the model, prioritizing them for human annotation and reducing redundant labeling of easy cases.
    
    \item[\textbf{Anomaly detection:}] identifying outliers or unusual patterns in text can help annotators focus on rare or unexpected cases, especially useful in large datasets where anomalies might carry significant insights.

    \item[\textbf{\textit{Automatic image and video tagging (for multimodal corpora):}}] if the corpus includes images or videos, tagging objects or actions automatically can assist annotators working on multimodal data by generating preliminary labels.
\end{itemize}

\noindent This list is far from comprehensive, but it underscores the wide range of opportunities in Kyrgyz NLP. In the next section, we propose our vision for the development of Kyrgyz NLP. However, there are many additional ways for independent contributions to progress in parallel.


\section{Roadmap for Kyrgyz NLP Development}\label{sec:roadmap}

\subsection{Proposed Milestones}

Developing Kyrgyz NLP requires a multifaceted approach that addresses the core linguistic challenges of the Kyrgyz language, as well as practical steps to ensure sustainability and growth. Below is a development roadmap broken down into specific steps, and then a timeline with three possible scenarios: the best case (with government support), the suboptimal case (minor or no support from the officials, and a certain involvement of the private sector), and the worst case (academic and community-based work without any external support whatsoever).

\begin{figure}[H]
    \selectlanguage{english}
    \centering
    \includegraphics[width=1\linewidth]{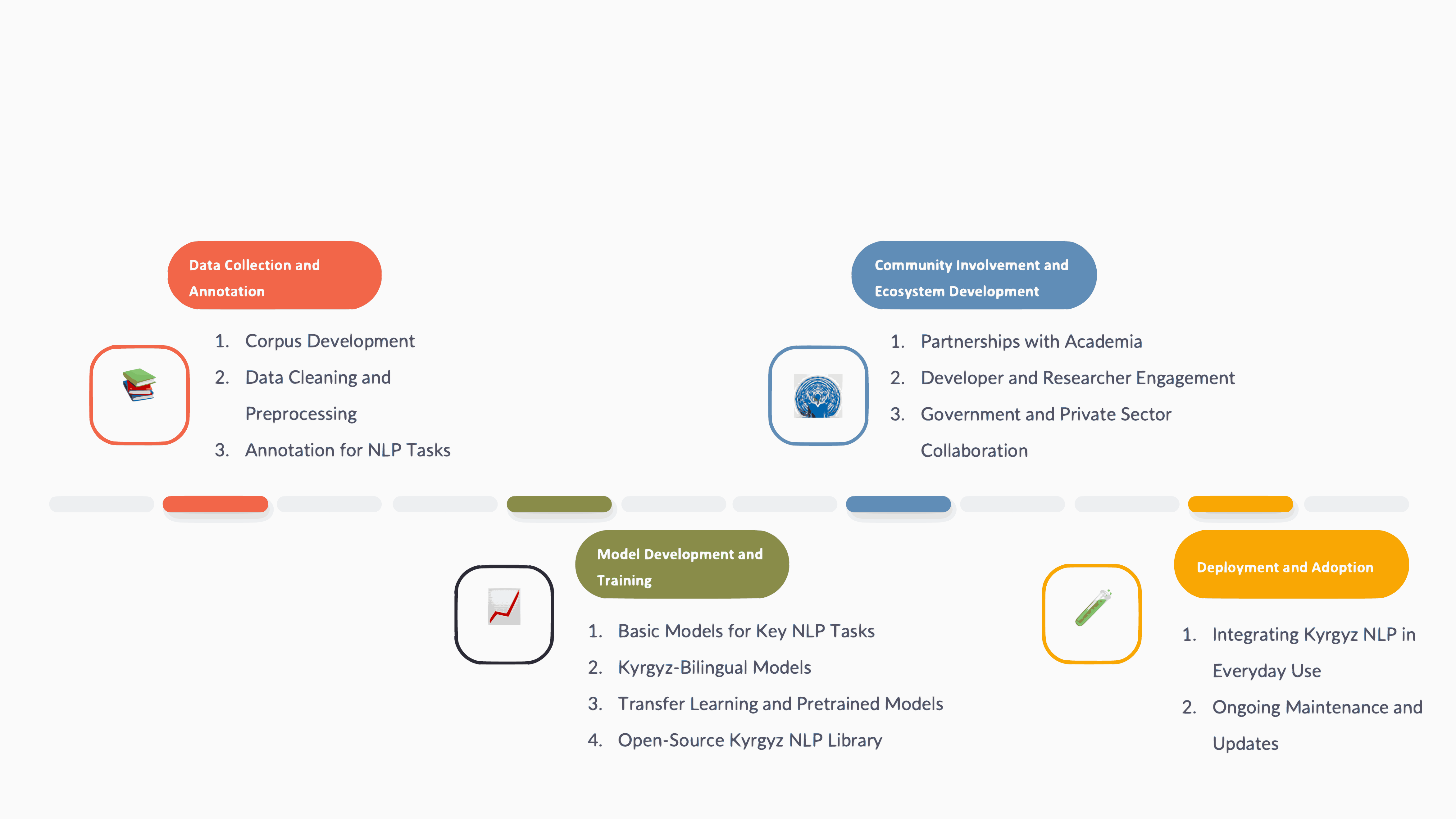}
    \caption{Milestones.}
    \label{fig:milestones}
\end{figure}

\subsubsection{Data Collection and Annotation}

\paragraph{Step 1: Raw Data Collection and Other Preliminaries.}
    
The foundation of Kyrgyz corpus linguistics and overall empirical language science lies in the creation of a comprehensive and diverse corpus, which should serve as the essential dataset for training and developing language models. The process must begin by gathering extensive datasets of Kyrgyz text from a variety of sources. These sources include literature, social media, news articles, academic publications, government documents, and other written materials that capture the full range of formal and informal language usage. This diversity is essential to ensure that the models represent the language accurately across different contexts, from casual conversation to formal communication.

Beyond raw data collection, it is critical to develop a robust suite of linguistic resources to support and expedite the annotation process. This includes digitized dictionaries, thesauri, and small pre-trained models that can streamline tasks such as syntactic analysis, morphological tagging, and part-of-speech labeling. These resources are foundational to achieving consistent, high-quality annotation. They also lay the groundwork for more efficient and scalable corpus processing, as models can rely on existing linguistic information to make quicker and more accurate decisions. Without these tools, creating a well-annotated corpus for Kyrgyz would be significantly slower and more prone to errors, which could impact the performance of the final language models.

To further enhance the corpus, spoken language should be incorporated alongside written text. This can be achieved by collecting and transcribing audio recordings from interviews, podcasts, radio broadcasts, and everyday conversational exchanges. Including both written and spoken language is crucial, as it allows the model to understand and process different styles, dialects, and nuances that occur in natural speech, which are often absent in written texts.

A robust corpus should also include metadata on regional dialects, language variations, and other sociolinguistic factors that can affect word choice, syntax, and tone. Additionally, steps must be taken to clean and preprocess the data, removing noise while preserving essential linguistic structures and variations. This process may involve filtering out irrelevant or low-quality content and structuring the data for easier processing by NLP algorithms.

Once a sufficiently large and representative corpus is compiled, not only can it be used for purely linguistic and documenting purposes (including updating the dictionaries, revising the grammar, historical linguistics, etc.), it can also be used to develop models that recognize, interpret, and generate Kyrgyz language with a high degree of accuracy, supporting downstream tasks such as machine translation, code switch detection, sentiment analysis, text generation, and many, many others.

\paragraph{Step 2: Data Cleaning and Preprocessing.}

Once the Kyrgyz corpus of unprocessed texts has been assembled, the next crucial step is to clean and preprocess the data, ensuring it is of high quality. This phase begins with systematically removing duplicates, irrelevant information, and any noise that may interfere with the accuracy of language models and linguistic analysis. Redundant text, unrelated content, and malformed entries need to be carefully filtered out to retain only the most relevant linguistic data.

After the initial cleanup, the text undergoes normalization, a process that aligns various forms of the language entities (e. g. words) into a consistent format. Given that Kyrgyz is spoken in diverse dialects and includes informal language variations, normalization is vital to ensure that the corpus represents standard forms that can be uniformly processed. This normalization also involves addressing differences in spelling, common informal expressions, and slang. Additionally, since Kyrgyz speakers often use different writing systems (such as Cyrillic, Perso-Arabic, or Latin), the data must be transliterated to a standardized script (however, preserving the original one). This step allows the model to manage variations due to transliteration or non-standard typing practices, commonly encountered on digital platforms and social media.

By the end of the data cleaning and preprocessing phase, the corpus is transformed into a well-structured, consistent, and high-quality dataset. These refined data provide a solid foundation for ensuring reliable linguistic analysis and training language models and other ML-based instruments, helping them understand the nuances and structures of Kyrgyz language while filtering out errors or inconsistencies that could negatively impact model performance.

\paragraph{Step 3: Annotation for NLP Tasks and Corpus Querying.}

With a clean and structured corpus, the next step is to annotate the dataset for specific NLP tasks, a critical process that allows language scientists to explore the laws and regularities of the Kyrgyz language and enables models to learn and perform essential linguistic functions. Annotation involves labeling text data to highlight parts of speech (PoS), build the syntax trees, identify named entities, determine sentiment, and mark other important linguistic features. Each of these tasks provides unique insights: POS tagging, for example, helps understand the grammatical roles of words within sentences, while Named Entity Recognition (NER) identifies names of people, places, organizations, and other key entities, facilitating more accurate comprehension of context. Sentiment analysis, in turn, adds emotional or opinion-based interpretation, which is valuable for tasks like analyzing public opinion, customer feedback, or the emotional variation patterns in prose and poetry.

To ensure that the annotation is accurate and reliable, guidelines must be created and strictly followed throughout the annotation process. These guidelines should specify rules for handling ambiguities, dialectal variations, and context-specific interpretations, offering annotators clear instructions on how to label each aspect of the text consistently. Additionally, if multiple annotators are working on the dataset, these guidelines help maintain uniformity, reducing variability and potential errors that could impact model training.

Once completed, the annotated corpus becomes a rich resource for Kyrgyz language studies as well as for training the machine learning models, enabling them to understand complex aspects of the Kyrgyz language such as syntax, context, and sentiment. This annotated dataset should support a wide range of NLP applications, from syntax parsing to semantic role labeling, from translation and summarization to question answering and conversational AI, providing the necessary foundation for sophisticated language processing capabilities and in-depth analysis.

\subsubsection{Model Development and Training.}

\paragraph{Step 1: Baseline Models for Key NLP Tasks.}

The initial focus should be on creating basic models to perform essential NLP tasks like part-of-speech (POS) tagging, Named Entity Recognition (NER), and machine translation, etc. These fundamental models form the backbone of more advanced applications and enable the system to recognize grammatical structures, identify key entities, and translate between Kyrgyz and other languages. By training on the annotated corpus, the models learn to accurately tag and classify words within sentences, paving the way for more complex language processing capabilities. These foundational models are crucial for developing language understanding and generation systems that can handle Kyrgyz-specific linguistic nuances.

\paragraph{Step 2: Kyrgyz Bilingual Models.}

To support cross-linguistic applications, the next step is to develop bilingual models that facilitate translation and communication between Kyrgyz and widely used languages like Russian and English. These models enable users to interact seamlessly in multiple languages and broaden the accessibility of Kyrgyz NLP applications. Developing Kyrgyz-Russian-Kyrgyz and Kyrgyz-English-Kyrgyz translation models is particularly important given the linguistic diversity in the region and the need for efficient communication across language barriers. By leveraging parallel datasets and aligning structures across languages, these bilingual models could serve as valuable tools for both general and specialized translation tasks.

\paragraph{Step 3: Transfer Learning and Pretrained Models.}

In this phase, transfer learning techniques are applied to leverage existing powerful models, such as BERT and GPT, which have already been trained on massive datasets. These models could be fine-tuned on Kyrgyz-specific datasets (as well as on raw data in related languages), allowing them to adapt to the unique characteristics of the Kyrgyz language without starting from scratch. Fine-tuning pre-trained models saves both time and resources, while also improving the quality and accuracy of Kyrgyz NLP tasks. This step enables the model to grasp specific Kyrgyz language patterns and vocabulary, enhancing its relevance and performance across different applications.

\paragraph{Step 4: Open-Source Kyrgyz NLP Library.}

The final stage in model development could be the creation of an open-source NLP library dedicated to Kyrgyz language processing. This library would compile a suite of ready-to-use models for tasks such as text classification, language generation, sentiment analysis, and more. Making this library open-source ensures that researchers, developers, and educators have easy and straightforward access to Kyrgyz NLP tools, fostering further innovation and collaboration. By offering a range of functionalities in one accessible platform, this library could empower users to build and expand upon Kyrgyz NLP applications, driving growth in both linguistic technology and cultural preservation.

\subsubsection{Community Involvement and Ecosystem Development.}  

\noindent Developing a sustainable and impactful Kyrgyz NLP ecosystem requires active community engagement and strong partnerships. This phase is dedicated to building a robust network involving academia, software developers, researchers, government entities, and the private sector, each contributing to advancing Kyrgyz NLP.

\paragraph{Step 1: Partnerships with Academia.} To establish a research-driven foundation for Kyrgyz NLP, partnerships with local universities and language institutions are essential. Collaborating with academia allows for the formation of specialized research teams focused on exploring and solving linguistic challenges unique to the Kyrgyz language. By engaging linguistics and computer science departments, these collaborations can provide students and faculty with hands-on experience in NLP research and development, creating a pipeline of skilled professionals. Joint projects with universities and language experts can also lead to innovations in language modeling, semantic analysis, and translation, further advancing the field of Kyrgyz NLP.

\paragraph{Step 2: Software Developers and Researchers Engagement.}
Encouraging active participation from developers and researchers accelerates progress in Kyrgyz NLP. This step includes organizing open challenges, hackathons, and funding initiatives that inspire developers and researchers to contribute their skills and ideas. One of the exemplary efforts is The Cramer Project, which organizes hackathons and datathons for developers and researchers. For example, they have hosted three datathons focused on advancing Kyrgyz NLP with support of High Technology Park of the Kyrgyz Republic, Ulutsoft, KSTU named after Iskhak Razzakov: MNIST\footnote{\url{https://thecramer.com/blog/datathon-1}}, morphology\footnote{\url{https://thecramer.com/blog/datathon-2}}, and NER\footnote{\url{https://thecramer.com/blog/datathon-3}}. These events provide participants with hands-on experience in solving real linguistic challenges and building practical tools for Kyrgyz language processing. 

By offering prizes, grants, and networking opportunities, these events make it possible for anyone interested in Kyrgyz NLP to participate in meaningful projects. Through hackathons, developers and students can work on specific NLP tasks, from sentiment analysis to NLU, while researchers may delve into more complex linguistic issues. This open collaboration not only fosters innovation but also creates a growing community of Kyrgyz NLP enthusiasts who can continue contributing to the ecosystem.

\paragraph{Step 3: Government and Private Sector Collaboration.}
Government and private sector support is critical to ensuring that Kyrgyz language technology receives adequate resources and visibility. Advocacy for government involvement can encourage the integration of Kyrgyz NLP applications into the public services and digital platforms (which is already being done, following the corresponding lawmaking initiatives). A digital-first approach allows government services to reach more citizens by communicating in Kyrgyz and sets a strong example for other organizations. In addition, collaboration with private companies is key to securing investments in Kyrgyz-language tech products, from smart devices to language-learning applications. By engaging the private sector, Kyrgyz NLP can become a viable business area, driving innovation in the industry.

This multi-pronged approach to community and ecosystem development helps ensure that Kyrgyz NLP is not only technically sound but also supported by a wide range of stakeholders, paving the way for sustainable growth and long-term impact.

\subsubsection{Deployment and Adoption.} 

\noindent The final phase focuses on bringing Kyrgyz NLP models to practical use and establishing a sustainable framework for ongoing development. Deployment and adoption are key to making Kyrgyz language technology accessible to the public, ensuring its relevance, and fostering long-term growth.

\paragraph{Step 1: Integrating Kyrgyz NLP in Everyday Use.}

To maximize impact, Kyrgyz NLP models should be integrated into public services and digital platforms used by the Kyrgyz-speaking community. This involves introducing Kyrgyz NLP tools such as translation services, virtual assistants, and e-governance applications that make everyday interactions and transactions more accessible in Kyrgyz. For instance, a Kyrgyz virtual assistant could streamline access to government services or provide localized information, while a reliable Kyrgyz translation tool could bridge language barriers in public communication. Beyond public services, promotion of the Kyrgyz language within popular tech spaces like social media platforms, messaging apps, and mobile applications can ease their everyday use. By embedding Kyrgyz language support into widely-used platforms, the language may become more accessible, modernized, and widely embraced within the digital ecosystem.

\paragraph{Step 2: Ongoing Maintenance and Updates}

For Kyrgyz NLP to remain effective and relevant, it requires continuous updates and maintenance. Language evolves over time, and so do the needs of its speakers, making it essential to keep NLP models aligned with these changes. A sustainable model for ongoing updates can be established by forming partnerships with research institutions, engaging the developer community, and securing funding from both public and private sectors. This model would allow for periodic improvements, expansion of the corpus with new data, and the retraining of models to account for emerging language trends, slang, or newly adopted terminology. By prioritizing consistent maintenance, Kyrgyz NLP tools can adapt to evolving linguistic and technological demands, ensuring they continue to serve users effectively and foster widespread adoption.

\subsection{Timeline}\label{ssec:timeline}

\begin{figure}[ht]
    \centering
    \includegraphics[width=1\linewidth]{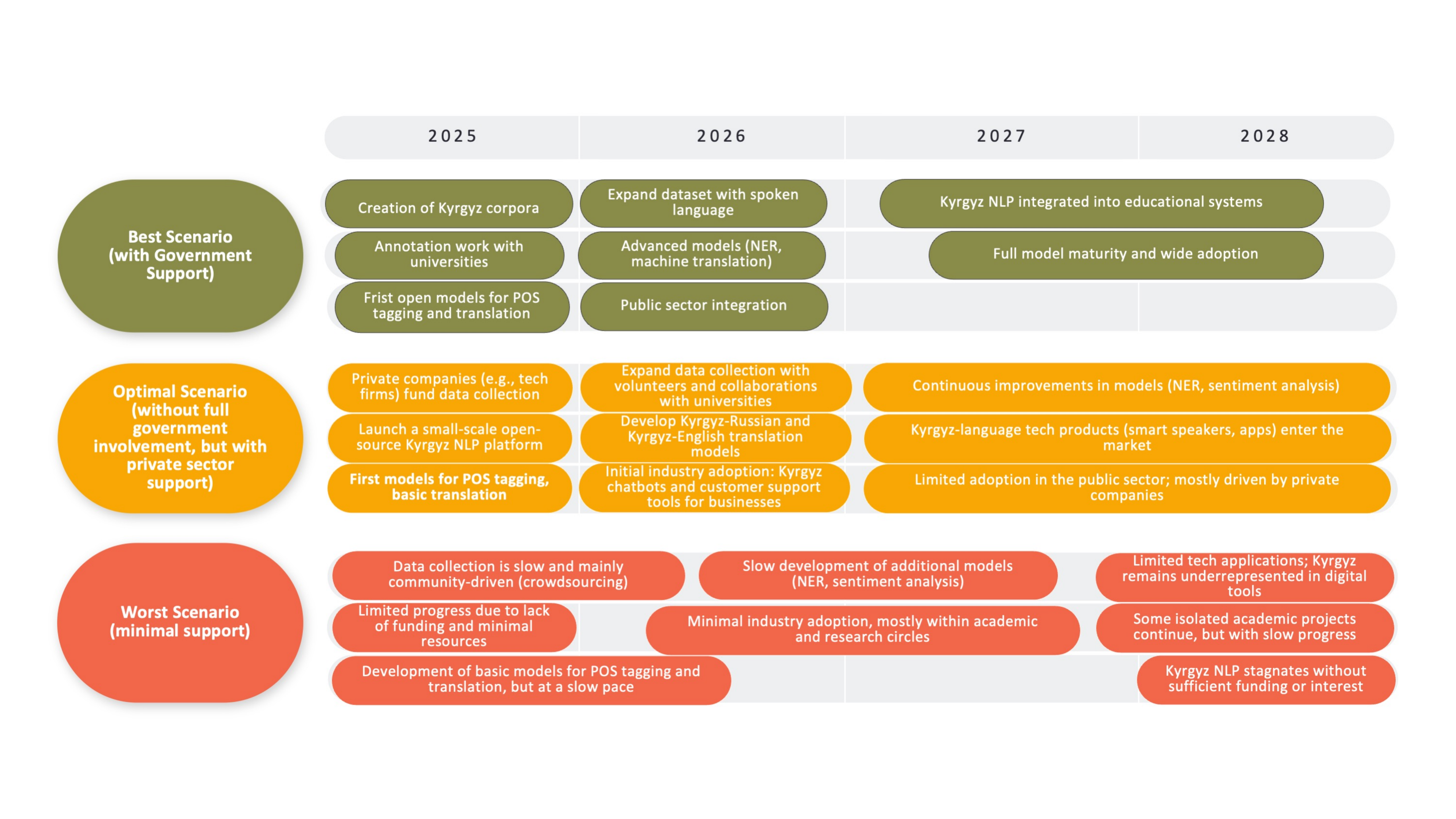}
    \caption{Timelines with scenarios.}
    \label{fig:timelines}
\end{figure}

The roadmap summarized in Fig.~\ref{fig:timelines} outlines key priorities and steps for developing Kyrgyz NLP, emphasizing how the involvement of various stakeholders (government, private sector, and academic community) can impact the speed and scope of overall progress in the field, as well as the adoptability of techniques of primary interest.

\subsubsection{The Best Scenario: Full Government Support}

\paragraph{Year 1}
\begin{itemize}
        \item Government endorses and funds Kyrgyz NLP development.
        \item Large-scale data collection from public libraries, public records, and media outlets (with all licensing questions solved).
        \item Partnerships with local universities for annotation work are established (and encouraged by the high authorities).
        \item First baseline models for POS tagging and translation are prepared --- together with reliable (large enough) benchmarks for evaluation.
        \item Support open-source Kyrgyz NLP initiatives: hackathons, challenges, conferences. 
\end{itemize}
    \paragraph{Year 2}
    \begin{itemize}
            \item The dataset collection process is expanded, integration of the speech data from the public and private sources is carried out.
            \item Development advanced models (NER, sentiment analysis, machine translation).
            \item The available models are integrated into the public sector applications (e. g., chatbots, e-governance).
            \item The software developers interested in the promotion of the Kyrgyz NLP engage in open-source products-related collaborations and hackathons.
    \end{itemize}
    \paragraph{Year 3-4}
    \begin{itemize}
            \item Continued government support ensures long-term sustainability.
            \item Kyrgyz NLP becomes a core part of the educational system in Kyrgyzstan.
            \item Kyrgyz tech products reach a certain level of maturity (popularity of the voice assistants, search engines, etc. with end users).
            \item Kyrgyz NLP models are widely adopted in the government and business sectors.
    \end{itemize}

\subsubsection{Suboptimal Scenario (without full government involvement, but with private sector support)}

\paragraph{Year 1}
\begin{itemize}
            \item Private companies (e. g., tech firms) fund data collection and initial model development.
            \item First models for POS tagging, basic translation, and text classification are released.
    \end{itemize}
\paragraph{Year 2}
    \begin{itemize}
            \item Data collection with volunteers' aid and collaborations with universities is done continuously.
            \item Kyrgyz-Russian and Kyrgyz-English translation models are developed and evaluated.
            \item Industry adoption starts: Kyrgyz chatbots and customer support tools for businesses.
    \end{itemize}
\paragraph{Year 3-4}
    \begin{itemize}
            \item Continuous improvements are introduced into the popular task models (NER, sentiment analysis).
            \item Kyrgyz-language tech products (smart speakers, apps) enter the market.
            \item Limited adoption in the public sector; mostly driven by private companies.
    \end{itemize}


\subsubsection{Worst Scenario (minimal support)}

\paragraph{Year 1}
\begin{itemize}
            \item Data collection is slow and mainly community-driven (crowdsourcing, student projects).
            \item The baseline models for POS tagging and translation are developed, but at a slow pace.
            \item The overall progress is limited due to the lack of funding and minimal human, data and computational resources.
 \end{itemize}       
    \paragraph{Year 2}
    \begin{itemize}
            \item The development of other models (NER, sentiment analysis) is slow.
            \item The results' industry adoption is minimal; mostly used within academic and research circles.
    \end{itemize}
    \paragraph{Year 3-4}
    \begin{itemize}
            \item Limited tech applications; Kyrgyz remains underrepresented in terms of digital tools.
            \item Some isolated academic projects do continue, but with slow progress.
            \item Kyrgyz NLP stagnates without sufficient funding or interest.
    \end{itemize}


\section{Conclusion}\label{sec:conclusion}

Advancing Kyrgyz NLP is crucial for ensuring the language's vitality in the digital age. Each step forward in developing NLP for less-resourced languages like Kyrgyz contributes to a more inclusive and linguistically diverse technological landscape, both now and in the future.

The advancement of Kyrgyz NLP presents both challenges and opportunities. Although the language faces significant obstacles due to resource limitations and linguistic complexity, there remains ample room for contributions from researchers, practitioners, and the community.

This overview may serve as a resource for researchers, developers, students, and policymakers engaged in Kyrgyz NLP and other less-resourced languages. By highlighting both progress and remaining challenges, we seek to inspire further research and development in this field.

The proposed roadmap emphasizes collaborative efforts across academia, industry, and government, focusing on developing larger, more diverse datasets, adapting state-of-the-art NLP techniques to Kyrgyz, and creating standardized evaluation benchmarks. Close collaboration among \textbf{all interested parties} is essential to overcoming current challenges.

We advocate for sustained efforts in data annotation, tool development, and model training, especially in the age of LLMs. The unique characteristics of each language require specialized approaches, and success with one language does not guarantee success with others.

By uniting our efforts, we can ensure that the Kyrgyz language is not left behind in the global advancement of NLP technologies. The preservation and development of Kyrgyz not only benefit its native speakers but also enrich the diversity of global linguistic resources.

\begin{credits}
\subsubsection{\ackname}
The work of A. Alekseev was supported by the Russian Science Foundation grant {\#23-11-00358}.
We express our sincere gratitude to the researchers, practitioners, and community members who have contributed to the development of Kyrgyz NLP resources and tools, as well as our collaborators and volunteers who supported data annotation and model development. Your commitment is essential in advancing the field and preserving the Kyrgyz language.
We extend special thanks to those whose enthusiasm and moral support have inspired our ongoing engagement with both earlier and recent developments in Kyrgyz NLP, forming the foundation of this paper.
We thank G. Dzhumalieva and A. Kasieva for their invaluable consultations on examples for this paper, and A. Tillabaeva for insightful discussions on corpus linguistics.
A. A. also acknowledges his advisor, G. Dzh. Kabaeva, for her trust and invaluable support throughout this endeavor, and S. I. Nikolenko for his assistance with proofreading the manuscript.
\end{credits}
\selectlanguage{english}
\bibliographystyle{splncs04}
\bibliography{references}
\end{document}